\documentclass[lettersize,journal]{IEEEtran}
\usepackage{amsmath,amsfonts}
\usepackage{algorithmic}
\usepackage{array}
\usepackage[caption=false,font=normalsize,labelfont=sf,textfont=sf]{subfig}
\usepackage{stfloats}
\usepackage{amsmath}
\usepackage{multirow}
\usepackage{pifont}
\usepackage[misc]{ifsym}
\usepackage{float}
\usepackage{amsfonts}
\usepackage{pifont}
\usepackage{cite}
\usepackage{booktabs}
\usepackage{caption}
\usepackage{graphicx}
\usepackage{colortbl}
\usepackage[table]{xcolor}
\usepackage{times}
\usepackage{epsfig}
\usepackage{graphicx}
\usepackage{adjustbox}
\usepackage{amsmath}
\usepackage{amssymb}
\usepackage{float} 
 \usepackage{amssymb}
\usepackage{stfloats}
\usepackage{textcomp}
\usepackage{stfloats}
\usepackage{url}
\usepackage{verbatim}
\usepackage{graphicx}
\usepackage{multirow}
\usepackage{hyperref}
\usepackage{fancyhdr}
\usepackage{threeparttable} 
\def\BibTeX{{\rm B\kern-.05em{\sc i\kern-.025em b}\kern-.08em
    T\kern-.1667em\lower.7ex\hbox{E}\kern-.125emX}}

\begin{document}

\title{Watch Where You Move: Region-aware Dynamic Aggregation and Excitation for Gait Recognition}

\author{
	\vskip 1em
	{
	Binyuan Huang{$\dagger$},
	Yongdong Luo{$\dagger$},
        Xianda Guo,
        Xiawu Zheng,
        Zheng Zhu,
        Jiahui Pan,
        Chengju Zhou        
	}

	\thanks{
		{This work was supported in part by the Major Projects of Colleges and Universities in Guangdong Province under grant 2023ZDZX2021 and the Guangdong Basic and Applied Basic Research Foundation under grant 2024A1515010524.
        
        B. Huang, J. Pan, C. Zhou are with the School of Artificial Intelligence, South China Normal University, Foshan 528225, China (email: 20192005065@m.scnu.edu.cn, panjiahui@m.scnu.edu.cn, cjzhou@scnu.edu.cn)

Y. Luo, X. Zheng are with Media Analytics and Computing Lab, Department of Artificial Intelligence, School of Informatics, Xiamen University, Xiamen 361000, China (email:  yongdongluo@stu.xmu.edu.cn,  zhengxiawu@xmu.edu.cn)

X. Guo is with the School of Computer Science, Wuhan University,
Wuhan 430079, China (e-mail: xianda\_guo@163.com)

Z. Zhu is with PhiGent Robotics, Beijing 100000, China (email: zhengzhu@ieee.org)

Corresponding author: Chengju Zhou.
$\dagger$ Joint first authors.

Digital Object Identifier 10.1109/TMM.2025.3613158.

Source code will be published at https://github.com/HUAFOR/GaitRDAE. 
		}
	}
}

\maketitle
\thispagestyle{fancy}
\fancyhf{} 
\renewcommand{\headrulewidth}{0pt}
\renewcommand{\footrulewidth}{0pt}
\fancyfoot[C]{%
  \scriptsize
 © 20xx IEEE. Personal use of this material is permitted. 
    Permission from IEEE must be obtained for all other uses, in any current or future media, 
    including reprinting/republishing this material for advertising or promotional purposes, 
    creating new collective works, for resale or redistribution to servers or lists, 
    or reuse of any copyrighted component of this work in other works.
}

\begin{abstract}

Deep learning-based gait recognition has achieved great success in various applications. The key to accurate gait recognition lies in considering the unique and diverse behavior patterns in \textbf{different motion regions}, especially when covariates affect visual appearance. 
However, existing methods typically use predefined regions for temporal modeling, with fixed or equivalent temporal scales assigned to different types of regions, which makes it difficult to model motion regions that change dynamically over time and adapt to their specific patterns.
To tackle this problem, we introduce a Region-aware Dynamic Aggregation and Excitation framework (GaitRDAE) that automatically searches for motion regions, assigns adaptive temporal scales and applies corresponding attention. Specifically, the framework includes two core modules: the Region-aware Dynamic Aggregation (RDA) module, which dynamically searches the optimal temporal receptive field for each region, and the Region-aware Dynamic Excitation (RDE) module, which emphasizes the learning of motion regions containing more stable behavior patterns while suppressing attention to static regions that are more susceptible to covariates.
Experimental results show that GaitRDAE achieves state-of-the-art performance on several benchmark datasets. 


\end{abstract}

\section{Introduction}
\IEEEPARstart{G}{ait recognition}, \textcolor{black} {aims to identify pedestrians by distinguishing their walking patterns.}
Currently, this technique reveals great application potential for video retrieval and security checks. 
According to the study in \cite{gaitori}, it is simple for humans to change their appearance or clothing, but more difficult to change their habitual movement and walking patterns. The fundamental advantage of gait stems from the fact that it compares human motion patterns rather than static patterns, making recognition less influenced by clothing color and texture. 
 \begin{figure}[t]
\begin{center}

\includegraphics[width=1.0\linewidth]{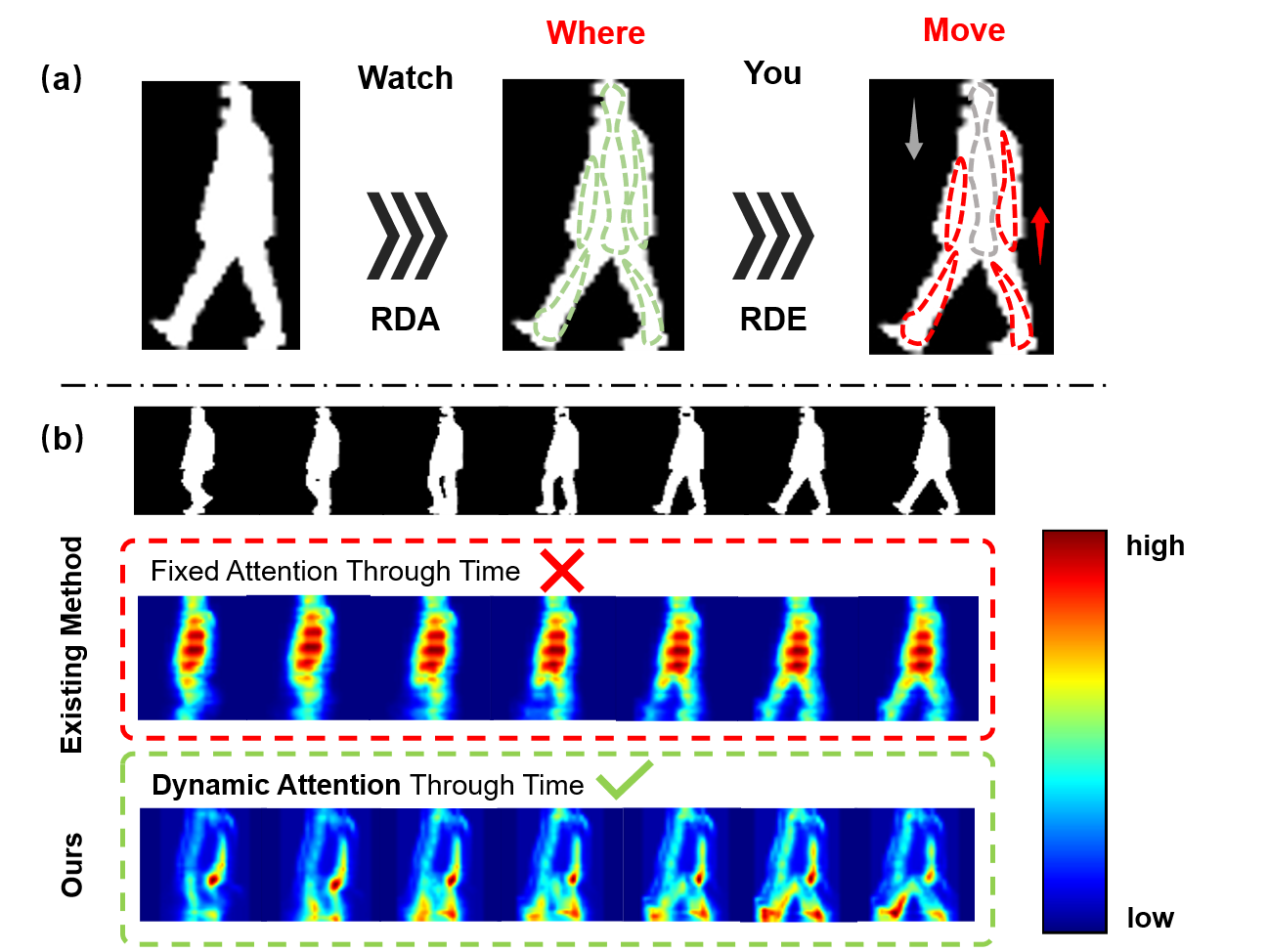}

\end{center}
   \caption{(a) An intuitive diagram of our methods. RDA is used to dynamically search the temporal receptive field of each region. RDE is utilized to adaptively enhance the attention of the motion region. 
   (b) Attention heatmaps from the existing method  and our method are illustrated. 
   The gradient bar to the right showcases how attention is variably allocated among diverse color segments.}
 \label{heatmap}
\end{figure}
While gait recognition methodologies have achieved notable progress, some issues remain within the existing techniques.
Firstly, 
the limitation of prior methods lies in their use of fixed or equal temporal receptive fields to model both motion and static regions, which hurt the model's adaptability. Motion regions, by their nature, exhibit more complex behavioral patterns, featuring longer periods and more intense variation frequencies. Consequently, deep understanding of such regions requires long-term temporal receptive fields. Conversely, static regions tend to have more homogeneous patterns, which can be captured using more restricted temporal receptive fields.
Secondly, 
how to dynamically excite the features of the motion region more comprehensively and deeply remains unresolved.
This conclusion is further supported by the visualization results 
of the existing method and our method in Figure~\ref{heatmap},
in which the network responds with higher weights to the static region, while those motion regions receive less response. As a result, these methods are more sensitive to changes in static regions, such as changes in shape caused by clothing and viewpoint variations. 
\textcolor{black}{
Ablation studies in \cite{gaitbase} further support this, indicating that the local feature extraction branches of part-based methods fail to perform effectively on outdoor datasets, resulting in performance degradation. The increased variability in clothing shapes and viewpoints in outdoor environments introduces additional challenges. If gait model is overly sensitive to static regions and fails to focus on more stable  motion patterns, it is inevitably more susceptible to performance drops due to interference from clothing and viewpoint changes.
}

In light of these limitations, it is critical to construct a region-aware dynamic gait model that can improve the network's understanding and focus on pedestrians' walking habits and dynamic patterns. Such a model would enable the adaptation of temporal receptive fields and attention based on each region's specific characteristics, thereby improving the model's robustness. Therefore, we propose a novel Region-aware Dynamic Aggregation and Excitation framework, called \textbf{GaitRDAE}. 
The method's critical concept is to automatically search for motion regions, assign adaptive temporal scales and apply the corresponding attention, thus enhancing the focus on discriminative patterns contained in motion regions and mitigating the influence of static regions susceptible to covariates.
Firstly, we present the Region-aware Dynamic Aggregation (\textbf{RDA}) module, 
in which the spatial-temporal convolution kernels are applied to the input feature to predict the required temporal receptive fields 
for each region and update the temporal receptive field according to the input samples. Furthermore, the Region-aware Dynamic Excitation (\textbf{RDE}) module is proposed to stimulate motion region feature learning in a comprehensive manner, including Spatial-wise Motion Excitation (\textbf{SME}) module and Channel-wise Motion Excitation (\textbf{CME}) module. 

Our research's key contributions are outlined as follows:
\begin{itemize}
    \item We propose the Region-aware Dynamic Aggregation (\textbf{RDA}) module to assign the optimal temporal receptive field for each gait region and adaptively adjust the region-aware temporal scale according to the input samples.
    \item We propose the Region-aware Dynamic Excitation (\textbf{RDE}) module to adaptively excite motion region features from spatial and channel dimensions simultaneously, which highlights the stable behavior patterns in the motion regions and suppresses changeable patterns within the static regions.
    \item Extensive experiments conducted on four popular datasets, i.e. GREW, Gait3D, CASIA-B, and OU-MVLP, demonstrate GaitRDAE achieves state-of-the-art performance, which  validates the robustness and adaptability of our proposed approach.
\end{itemize}

The rest of the paper is organized as follows:
Section II provides an overview of the technical background in gait recognition. 
Section III delves into the specifics of the RDA and RDE modules, the core components proposed in this study. 
Section IV outlines the datasets utilized in our experimental work, offers a performance comparison with  state-of-the-art methods across several popular datasets, and includes analyses of ablation studies as well as the results of visualizations.
Finally, Section V offers a comprehensive summary of our framework as presented in this paper.

\section{Related Work}
\subsection{Gait Recognition} 
In the field of gait recognition, two principal methodologies are recognized: model-based and appearance-based approaches. 
\subsubsection{Model-based methods} 
Model-based approaches leverage pose information obtained from pose estimation methods to model walking patterns\cite{posegait, 
gaitgraph,
TMM3,
TMM1}, and recently these techniques are receiving increasing attention\cite{modelbased20-GaitPT,modelbased19-fastpose,modelbased23-mom,TMM5,modelbased18-huang } due to the successful development of pose estimation and skeleton-based action recognition techniques \cite{sklten1,sklten2,sklten3}.
For example, 
PoseGait\cite{posegait} utilizes compact 3D human body pose features and effectively employs CNN for extracting gait features, thereby improving recognition performance.

CAG \cite{modelbased18-huang} adjusts dynamically according to the unique characteristics of individual skeleton sequences and their associated viewing angles.
\textcolor{black}{
Gait-TR\cite{gait-tr} pioneers the application of the spatial transformer framework in skeleton-based gait recognition, demonstrating superior capability in gait feature extraction from the human skeleton compared with the prevalent graph convolutional network.
GPGait\cite{gpgait} improves the generalization of pose-based gait recognition across datasets by using a Human-Oriented Transformation (HOT) to achieve a unified pose representation, a Human-Oriented Descriptors (HOD) for discriminative multi-features, and a Part-Aware Graph Convolutional Network (PAGCN) for efficient graph partitioning and local-global spatial feature extraction.
}
Although these methods produce gait representations that are robust to the effects of covariates like changing clothing and camera viewpoint, they suffer from issues such as a lack of identity-related shape features. 
\subsubsection{Appearance-based methods} 
Appearance-based methods can be further classified into three types: template-based methods, methods utilizing unordered sets, and those employing ordered sets.

In template-based techniques,
a common practice in gait recognition involves distilling gait sequences into singular representative images\cite{gei2,plus1-gei,gei-xu, plus6-gei,geinet }. A notable template in this context is the Gait Energy Image (GEI), which uses temporal averaging to reduce computation. Researchers such as Wu et al.\cite{plus6-gei} and Shiraga et al.\cite{geinet} apply CNN-based methods to extract critical gait features from GEIs.
For learning view-specific feature representations, He et al.\cite{gei2} utilize generative adversarial networks. In cross-view gait recognition, Xu et al.\cite{gei-xu} pioneer an approach that combines coupled projections with multi-view subdomain learning, efficiently representing gait's spatial-temporal elements.
Nevertheless, the Gait Energy Image (GEI) method adversely affected gait recognition accuracy by overly compressing the subject silhouettes across the full temporal sequence, resulting in a loss of critical temporal information.

In contrast, unordered set-based methodologies have taken a different approach\cite{gaitset, gaitgln,gaitbase,srn } . GaitSet\cite{gaitset}, for instance, posits that a silhouette's visual appearance inherently encapsulates its temporal positional information. Consequently, gait is treated as a set, enabling the potential reconstruction of temporal cues. Building upon this concept, GLN\cite{gaitgln} enhances the discriminative power of gait representation by incorporating feature pyramids into deep neural networks.
On another front, to further enhance discriminative capabilities,

and GaitBase\cite{gaitbase} establishes a powerful and concise ResNet-like model structure, using an unordered set as input.
However, they do not  explicitly capture fine-grained details of gait motion, potentially missing out on the ability to model intricate motion nuances.
Moving into the domain of ordered sequence-based techniques\cite{  
lagrangegait, 
mt3d, 
gaitmask,
gaitmstp,
gaitctcg,
gaitstrip,
TMM2,
hid2021,
TMM4,
gaitage
}:
MT3D \cite{mt3d} employs a multi-branch 3DCNN architecture designed to capture frame-level features across various temporal scales.  
GaitStrip \cite{gaitstrip} enhances gait recognition by combining spatial-temporal, frame-level, and strip-based features within a multi-level framework.
L-Gait \cite{lagrangegait} develops both a module for extracting second-order motion and a compact module for view embedding,
which are intended to address the fact that current cross-view approaches do not explicitly take into account the view-specific features.
In the context of our current study, our methodology aligns with this ordered sequence-based category. 
Specifically, we take the silhouette of the subject as our input, and our proposed GaitRDAE is designed to operate within this framework.

\subsection{Temporal Modeling} 
The landscape of existing temporal modeling techniques in gait recognition can broadly be delineated into three primary categories based on their region-modeling strategy: Global-aware temporal learning, Local-aware temporal learning, and the Global-Local temporal learning. 
Regarding the Global-aware paradigm, methodologies such as GaitSet\cite{gaitset}, GLN\cite{gaitgln}, and MT3D\cite{mt3d}, GaitBase\cite{gaitbase} regard each gait frame as an integrated entity. To aggregate temporal information, these models typically employ strategies like temporal pooling or take advantage of 3D convolution.
In the domain of the Local-aware model, 
GaitPart\cite{gaitpart} and CSTL\cite{cstl} equally divides the input feature maps into several strips horizontally and extracted temporal features for each strip independently. 
For the last type, 
GaitGL\cite{gaitgl} employs 3D convolution kernels for extracting and aggregating gait characteristics from the global region and each local region that is divided horizontally and uniformly. 
3DLocal \cite{3dlocal}, establishes six predefined local paths, each corresponding to a specific local region, 
and utilizes 3D convolution kernel's capacity to derive spatiotemporal information
from the global region and the six local regions.
However, these above methods introduce a prior human division of gait regions, which could not accurately localize motion regions that change dynamically over time. Also, due to the change of walking posture and capture view in different datasets\cite{gait3d,grew}, the predefined region needs to change accordingly.

\textcolor{black}{
Recently, parsing-based methods\cite{parsing-mm1-allyouneed,parsing-mm2-allyouneed,parsing-tmm} have been proposed to address this issue by providing a more dynamic and anatomically accurate representation of human parts. For instance, 
Parsing-Gait\cite{parsing-mm2-allyouneed} introduces the Gait Parsing Sequence (GPS), which uses fine-grained human segmentation to encode shapes and dynamics of body parts during walking, significantly improving the representation's information entropy compared to traditional silhouette-based methods. 
Similarly, LandmarkGait \cite{parsing-mm1-allyouneed}emphasizes the use of intrinsic human parsing to enhance the accuracy of gait recognition by focusing on stable and distinctive human body landmarks, addressing issues related to pose variations and occlusions. 
GaitParsing\cite{parsing-tmm} highlights the advantages of parsing-based methods in providing detailed and robust representations of human parts, which can significantly enhance gait recognition accuracy in complex scenarios.
These parsing-based approaches offer several advantages over traditional methods. They provide higher information entropy at the pixel level, enabling the encoding of more detailed and discriminative features of human motion. This high granularity allows for better handling of variations in pose, clothing, and occlusions, making the models more robust in real-world applications.
Both parsing-based methods and our proposed GaitRDAE framework share a common goal of improving gait recognition by focusing on the distinct characteristics of different body regions. However, parsing-based methods achieve this by using detailed human segmentation to provide anatomically accurate representations, while our GaitRDAE introduces additional innovations that enhance dynamic region-aware temporal modeling.
Unlike existing methods which use equal or fixed temporal receptive fields to learn patterns of different regions,  the proposed Region-aware Dynamic Aggregation (RDA) module assigns adaptive temporal scales based on the unique motion characteristics of each region, which enhances GaitRDAE's ability to capture diverse and dynamic gait patterns, providing a significant advancement in handling temporal variations and improving robustness in gait recognition tasks.
}

\subsection{Motion Modeling} 
In this paper, we claim that motion features are local inter-frame high-frequency variations that are separated from static representations\cite{smt}. Different from the spatial-temporal features that focuses on extracting temporal low-frequency information (the global representation of the aggregated frames), the motion features focus on temporal high-frequency information (the motion edges and regions of the aggregated frames) and suppress the static features. 
GaitMotion\cite{gaitmotion} and SM-Prod\cite{SM-Prod} are combined with optical flow motion information for gait recognition. However, computing optical flow requires huge computation costs and cannot be taken to good effect for practical applications. 
In recent, 
L-Gait\cite{lagrangegait} develops an innovative module for extracting second-order motion, which operates on the principle of computing self-similarity between frames,
but the extracted motion features are still mixed with information from static regions and are therefore still susceptible to static covariances. 
DyGait \cite{dygait} generates a motion feature map by calculating the difference between the gait characteristics of each individual frame and a standard gait template, which is derived from an average function.
However, how to model motion features more comprehensively and deeply has not been addressed.
To address this problem, we propose Region-aware Dynamic Excitation (RDE) module 
that effectively utilizes both channel-wise and spatial-wise aspects
in a flexible manner.

\section{Method}
\begin{figure*}[t]
\begin{center}
\includegraphics[width=0.9 \linewidth]
{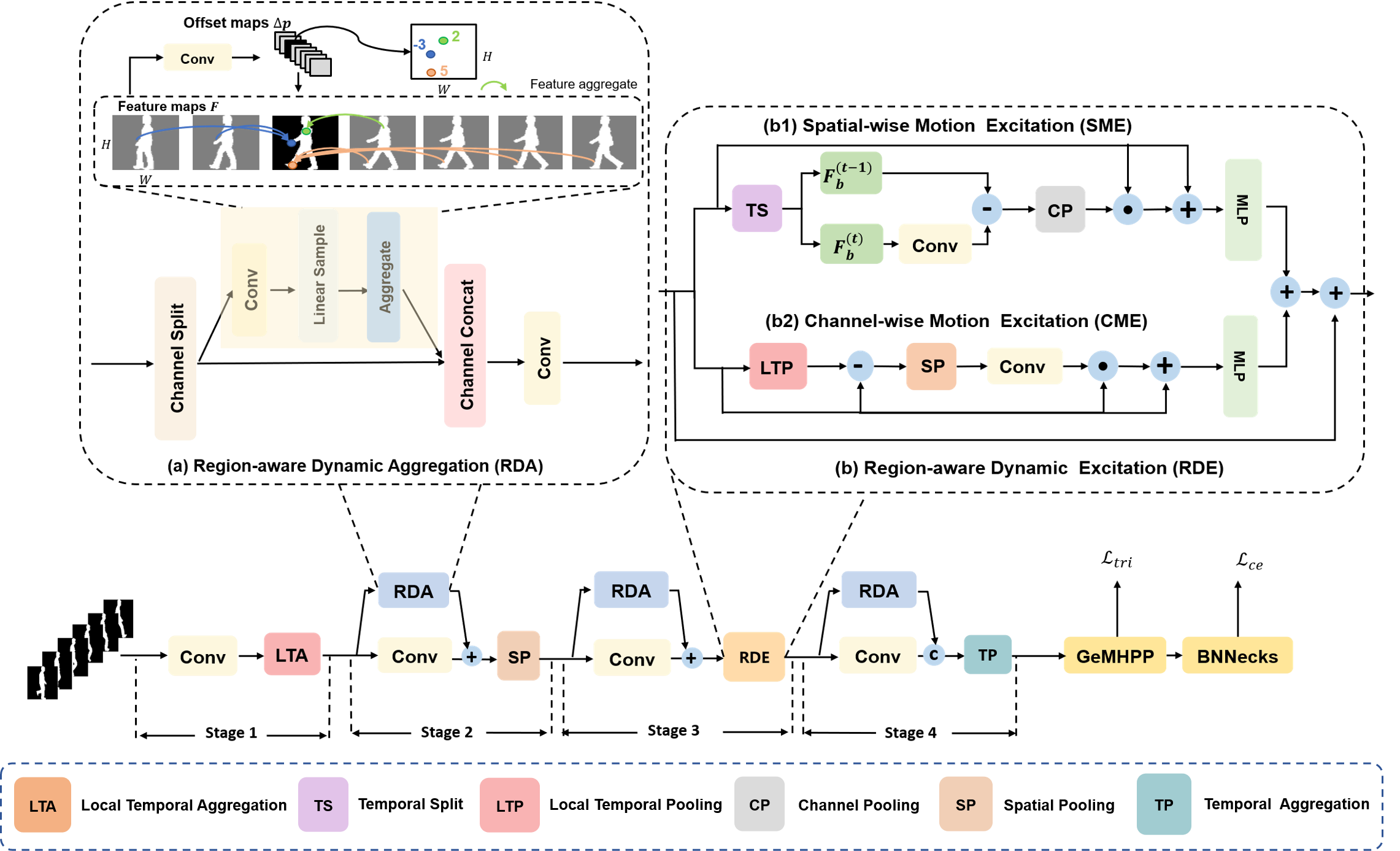}
\end{center}
\captionsetup{font= normalsize}
   \caption{
   An overview of the GaitRDAE architecture is presented. Our system processes a sequence of T gait frames as its input. The term "Conv" represents the convolutional layers within the framework. Feature fusion in the model employs element-wise addition, denoted by “+”, and height-level concatenation, represented by “C”. The loss functions used in this model are the triplet loss ($L_{tri}$) and the cross-entropy loss ($L_{ce}$).
   }
 \label{overview}
 \label{overview}
\end{figure*}

The structure of GaitRDAE is shown in Figure~\ref{overview}. The input gait features $F_{in}\in R^{T\times C\times H\times W}$ will go through four processing stages.  $T$, $C$, $H$, and $W$ symbolize the  sequence length, number of channels, spatial height, and width, respectively. 
In \textbf{Stage 1}, after a initial convolutional module, a Local Temporal Aggregation (LTA) \cite{gaitgl} module  subsequently captures local temporal information  of the features, resulting in \( F_{s1} \in \mathbb{R}^{T/3 \times C \times H \times W} \). 
\textcolor{black}{
LTA\cite{gaitgl} is designed to aggregate temporal information while preserving spatial details, integrating temporal data from local clips and maintaining spatial information. Then, RDA is designed to dynamically search for critical motion regions and assign adaptive temporal scales for different regions.
} 
To note that in the following three stages, features are all first processed by parallel RDA and convolutional modules, and then fed to a post-process module. In \textbf{Stage 2}, RDA is followed by a spatial pooling, yielding \( F_{s2} \in \mathbb{R}^{T/3 \times C \times H/2 \times W/2} \). 
In \textbf{Stage 3}, RDA is followed by a RDE module, yielding \( F_{s3} \in \mathbb{R}^{T/3 \times C \times H/2 \times W/2} \). 
And in \textbf{Stage 4}, RDA is followed by a temporal maxpooling function to generate \( F_{s4} \in \mathbb{R}^{C \times H \times W/2} \).
\textcolor{black}{
It is crucial to position the RDA module after RDE, which allows the RDA to allocate spatiotemporal receptive fields more precisely by leveraging RDE-activated features. 
This arrangement enhances the model's ability to capture complex motion patterns by providing a more refined analysis of motion characteristics.
}
\textcolor{black}{The placement of RDA and RDE modules is strategic: the RDA modules are inserted from Stage 2 onwards because shallow features in Stage 1 lack sufficient temporal-spatial encoding necessary for effective dynamic aggregation. By incorporating RDA in the later stages, the network can progressively learn dynamic temporal-spatial characteristics across different feature levels. Specifically, in Stages 2 to 4, different dimensions of information are aggregated sequentially: Stage 2 aggregates spatial features using Spatial Pooling (SP), Stage 3 employs the RDE module to apply attention to global motion features, and Stage 4 uses Temporal Pooling (TP) to aggregate temporal features. This order ensures that after spatial features are aggregated in Stage 2, local spatial information is enhanced, and before aggregating temporal features in Stage 4, the temporal information is sufficiently enriched, thereby effectively extracting motion features.}
After all the four stages, the Generalized-Mean Horizontal Pyramid Pooling\cite{gaitgl} are further utilized to extract part-level features $F_{gem}\in R^{C\times H}$. 
Finally, the BNNecks\cite{bnnecks} are adopted to perform feature space restructuring. 
In the following of this section, the detailed structure of RDA and RDE, as well as the overall optimization strategy of GaitRDAE, will be introduced sequentially.

\subsection{Region-aware Dynamic Aggregation (RDA)}

Previous approaches used predefined regions
as well as fixed and unified temporal scales for temporal modeling.
In contrast, our proposed RDA dynamically searches distinct temporal scales for different  regions based on their unique motion characteristics.
Features are aggregated  along temporal dimension to achieve adaptive temporal scale updates, and the temporal range of the aggregation is determined by learnable temporal offset value for each individual space-time pixel area. 
As shown in Figure \ref{overview}(a), 
to preserve spatial modeling capability of features while enabling effective temporal modeling,
only a portion, with ratio $r \left( 0<r<1 \right)$ of feature channels are  selected to 
obtain dynamic temporal receptive field for different regions. A 3D CNN with multiple input channels and single output channel is adopted to aggregate comprehensive dependencies between temporal, spatial, and feature channel dimensions and generate the required space-time pixel-wise temporal offset. \textcolor{black}{Such convolutional operations enforce spatial smoothness in the predicted offsets, causing pixels belonging to the same semantic region (e.g., leg area) to receive similar temporal offsets without appearing significantly different offsets.} To expand the flexible temporal range to the whole video length, the tanh activation function is adopted and  the activated value is multiplied by $T$:
\begin{equation}
\Delta t=T\odot tanh \left(K_{t,h,w}^{ST1} \circledast F \right)
\end{equation}

where $\Delta t\in \mathbb{R}^{T\times 1\times H\times W}$represents the learnable temporal offset value, 
and $K_{t,h,w}^{ST1}$ denotes the  spatio-temporal convolution filter with kernel size 3$\times$3$\times$3.
The learnable offset matrix $\Delta t$ is used for $ C\times r $ channels to  calculate.
Since the offset value $\Delta t$ is typically fractional, inspired by \cite{deformable}, bilinear interpolation is adopted to obtain  feature value:
\begin{equation}
\begin{aligned}
 O_{c,t,h,w}&= \left(\lceil \Delta t \rceil-\Delta t \right)\odot F_{c,t+\lfloor \Delta t \rfloor,h,w}\\
            &+\left(\Delta t - \lfloor \Delta t \rfloor \right) \odot F_{c,t+\lceil \Delta t \rceil,h,w}   
\end{aligned}
\end{equation}

Then, a time dimensional average pooling within the temporal range is calculated to aggregate pixel feature sequences from $t$ to $t+\Delta t$ so as to obtain the adaptive temporal receptive field for each pixel (Note that the value of $t+\Delta t$ is clamped at $[0, T-1]$ to ensure that it is meaningful): 

\begin{equation}
F_{adaptive}= \frac{1}{ \lceil \Delta t \rceil +1} \left( O_{c,t,h,w} + \sum_{i=t}^{t+ \lfloor \Delta t \rfloor }F_{c,i,h,w} \right)
\end{equation}
where $ F_{adaptive}\in \mathbb{R}^{T\times \left(C\times r\right) \times H\times W}$represents the aggregated  feature representation. 
After that, 
the temporal aggregated feature frames of each selected channels are concatenated with the remaining unprocessed feature in the channel dimension to recover the original shape $\mathbb{R}^{T\times C\times H\times W}$. 
An additional learnable spatio-temporal convolutional kernel is used to facilitate the interaction between adaptively temporal-aggregated feature and unprocessed spatial features, effectively combining and enhancing dynamic spatio-temporal relationships \textcolor{black}{and helping maintain spatial consistency while capturing region-specific temporal dynamics}:
\begin{equation}
F_{RDA}= K_{t,h,w}^{ST2} \circledast 
[ F_{adaptive}; F_{unprocessed}]
\end{equation}

\begin{figure}[t]
\begin{center}

\includegraphics[width=1.0 \linewidth]{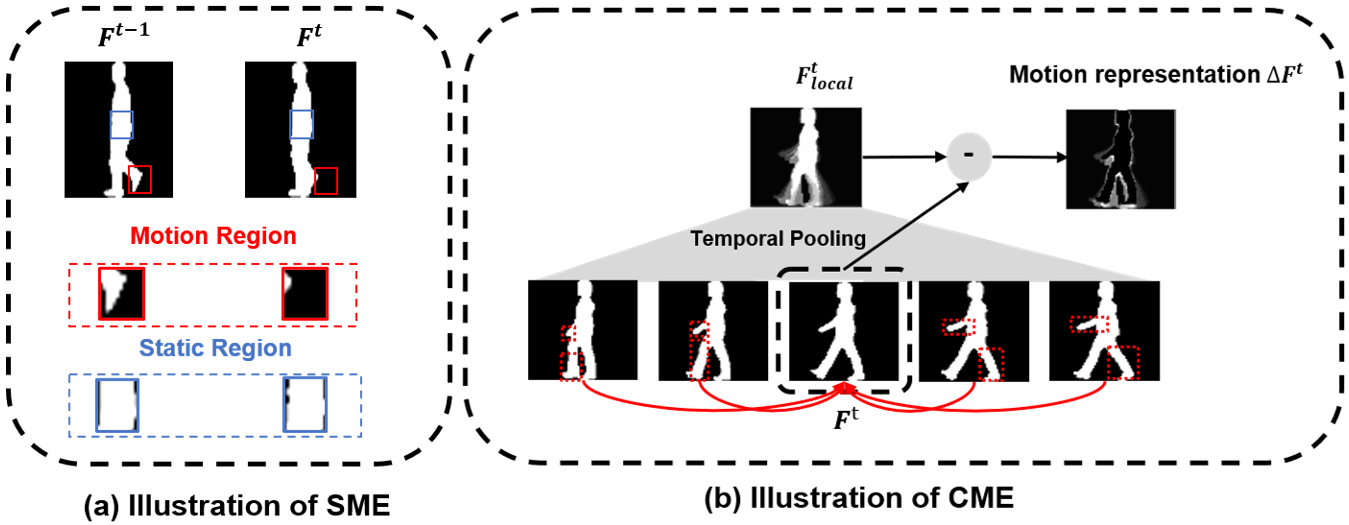}
\end{center}
   \caption{An intuitive illustration of Spatial-wise Motion Excitation (SME) and Channel-wise Motion Excitation (CME) .}
   \captionsetup{font=large}
   \label{sme_cme}
\end{figure}
\subsection{Region-aware Dynamic Excitation (RDE)}
RDE is proposed to dynamically excite motion features from spatial and channel dimensions. RDE contains two major submodules: Spatial-wise Motion Excitation (SME) and Channel-wise Motion Excitation (CME),
In RDE, features are processed parallelly by SME and CME, and the outcome of these two modules are summed up together with the input feature to generate the  output of RDE, as shown in Figure 2(b).
\\
\subsubsection{Spatial-wise Motion Excitation (SME)}
As shown in Figure~\ref{sme_cme}(a), motion regions have larger semantic changes (response differences) than static regions.
Based on this finding, SME assigns higher spatial attention to motion regions and lower focus to static regions by converting the response difference between adjacent frames into spatial attention weights.
Figure~\ref{overview}(b1) depicts structure of SME,
after time splitting, a copy of input feature sequence is convolved with a channel-wise 2D convolution kernels for inter-frame feature alignment \cite{smt}, and a set of channel-wise spatial weight masks are then generated for different regions by subtracting adjacent frames between convolved and unconvolved sequences. The final spatial  mask is a channel aggregation of all the weight mask:
\begin{equation}
 M_{SME} =\frac{1}{C} \sum_{i=1}^{C} [K_{h,w}^{SME} \circledast F_{t} - F_{t-1}]
\end{equation}
where $M_{SME} \in \mathbb{R}^{T \times 1\times H\times W}$ denotes the  spatial region mask generated based on the response difference of adjacent frames and its first frame is a copy of the second frame to settle the boundary issue. 
$K_{h,w}^{SME}$ is the 2D convolution kernel with size 3$\times$3. 

Afterward, the unified spatial weights are first filtered with a sigmoid function and then dot multiplied with all the channels of the input feature sequences 
to emphasize the discriminative motion regions and suppress features from static regions. 
\textcolor{black}{
A residual connection across the weighting steps is
introduced to help convergence and ensures no critical information is lost while emphasizing motion regions and semantics. 
} 
Finally, a multilayer perceptron is utilized to provide further  wider range of pattern modeling:

\begin{equation}
F_{SME}= MLP\left(\sigma\left(M_{SME}\ \right)\ \odot\ F\ + F\right)
\end{equation}  
where the sigmoid function is represented by $\sigma$, and the SME's final output is 
$F_{SME} \in\mathbb{R}^{T\times C\times H\times W} $.

\subsubsection{Channel-wise Motion Excitation (CME)}
It’s known that the feature maps of different channels depict different semantic patterns. Some channels focus on depicting the semantics of static regions, while others focus on the semantics of motion regions \cite{tea}. 
As previously mentioned, the semantic properties of motion regions are crucial in recognizing gait patterns. Consequently, CME is introduced to enhance the channels associated with motion. The architecture of CME is depicted in Figure~\ref{overview}(b2). 
Unlike SME, which is primarily concerned with dynamic activation in the spatial domain,  CME concentrates on channel dimension and pays more attention to motion variations in local temporal contexts.
A local temporal average pooling is first utilized to aggregate temporal contexts, which is formulated as:

\begin{equation}
F_{local}=\ P_{local} \circledast F 
\end{equation}
where $F_{local}{\in\mathbb{R}}^{T\times C\times H\times W}$  denotes the local temporal representation and $P_{local}$ denotes the local temporal average pooling kernel. 
The absolute difference between input feature and its low-frequency filtered version of $F_{local}$ is calculated to server as motion representation:

\begin{equation}
\begin{aligned}
\Delta{F}^t=& |F^t-F_{local}^t| 
\end{aligned}
\end{equation} 
where $\Delta{F}^{t}$ denotes the high-frequency motion information separated from static representation at frame t.
As shown in Figure \ref{sme_cme}(b), each representation  
$\Delta{F}^{t}$    aggregates the high-frequency variation of $k$ nearest neighbors.

This paradigm of local motion encoding allows our model to focus on high-frequency information and dynamic patterns, thus giving more focus to habitual underlying behavior patterns than changeable extrinsic static representations. To further excite the motion-sensitive channels, 
CME incorporates a global spatial pooling to aggregate the spatial features:  
\begin{equation}
\Delta F^{A}\ =\frac{1}{H\times W} \sum_{i=1}^{H} \sum_{j=1}^{W} \Delta{F[:,:,i,j]}, \Delta F^{A} \in \mathbb{R}^{T\times C\times 1\times 1} 
\end{equation}

Following this, a temporal convolution kernel are adopted to excite motion-sensitive channels and further enhance the channel-temporal relationship:
\begin{equation}
M_{CME} = K_{t}^{CME} \circledast \Delta F^{A}
\end{equation}
where $K_{t}^{CME}$ represents the  temporal convolution kernel with kernel size $3\times1\times1$,
and $M_{CME}$ is the output channel weights.
Finally, similar to SME, the channel attention weights $M_{CME}$ is activated by a sigmoid function and multiplied with the input features $F$, 
after a residual concatenation and the final feature mapping of MLP, the feature space is adjusted as:
\begin{equation}
F_{CME}= MLP \left(  \ \sigma\left(M_{CME}\ \right) \odot F+ F\right)
\end{equation}
where the sigmoid function is represented by $\sigma$, and the CME's final output is 
$F_{CME} \in\mathbb{R}^{T\times C\times H\times W} $.

\subsection{Optimization Function}
In line with the optimization framework detailed in \cite{gaitbase,gaitgl,cstl}, our training model utilizes a dual-loss strategy, incorporating both triplet loss ($L_{tri}$) \cite{triplet} and cross-entropy loss ($L_{cse}$), applied independently to each lateral feature slice.
The triplet loss, which functions as an adjunct to cross-entropy loss, aims to widen the distance between different classes while narrowing the distance within the same class. Given a set of three related gait sequences $(A,P,N)$, where $A$ represents the anchor, $P$ represents the positive instance, and $N$ represents the negative instance, the triplet loss is defined as:
\begin{equation}
\label{tri}
    L_{tri}=[m+d(A-P)-d(A-N)]_+
    \end{equation}
where $d(\cdot)$ denotes the Euclidean distance and $m$ denotes a predefined margin.

The cumulative loss is thus defined as:
$ L_{combined} = L_{tri} + L_{cse} $.

\section{Experiments}
\subsection{Datasets}
In this paper, we conduct experiments on two wild datasets, GREW\cite{grew,grew2} and Gait3D\cite{gait3d} 
together with two indoor datasets, CASIA-B\cite{casia-b} and OU-MVLP\cite{ou-mvlp}.

\textbf{GREW}  is a comprehensive gait dataset from public spaces, featuring 128,671 instances from 26,345 individuals recorded by 882 cameras. It includes diverse scenarios like obstructions and various carrying conditions, offering a rich set of variables for realistic gait analysis. The dataset is split for model training (102,887 instances for 20,000 individuals), validation (1,784 instances for 345 individuals), and testing (24,000 instances for 6,000 individuals). In the test phase, the initial two instances per individual constitute the gallery set and the subsequent instances form the probe set.

\textbf{Gait3D} represents a voluminous dataset of 3D gait patterns captured in a vast commercial retail environment, with 25,309 instances of 4,000 individuals captured by 39 cameras. This dataset captures complex pedestrian dynamics, including non-uniform speeds and paths, enhancing the challenge for gait recognition systems. It's constructed for development and evaluation, assigning 3,000 individuals to training and 1,000 to testing, where one instance per subject forms the gallery set and the rest as the probe set.

\textbf{CASIA-B}, a comprehensive indoor gait dataset, is recognized for its extensive cross-angle analysis, encompassing recordings of 124 individuals from 11 distinct angles (ranging
from 0° to 180° at intervals of 18°). 
The dataset captures three gait types: normal walking (NM, 6 sequences), with a bag (BG, 2 sequences), and in a coat (CL, 2 sequences). It uses 74 subjects for training and 50 for evaluation. The first four NM sequences are the gallery set, while the last 2 NM, 2 BG, and 2 CL sequences form the probe set for evaluation.

\textbf{OU-MVLP}, a large indoor gait dataset with 10,307 individuals recorded from 14 angles between 0°-90° and 180°-270°, at 15° intervals. It includes two sequence types ('00' and '01') for each angle. The dataset splits 5,153 subjects for training and 5,154 for evaluation. In evaluations, '01' sequences form the gallery set, and '00' sequences serve as the probe set.

\begin{table*}[t]
\small
\renewcommand{\arraystretch}{1.25}
\setlength{\tabcolsep}{5.3mm}
\centering
\caption{Performance comparison of Rank-1 (\%), Rank-5 (\%), mAP(\%)  on Gait3D.}
\begin{tabular}{c|c|ccc|ccc}
\toprule

Methods                      			& Source & Rank-1  & Rank-5  & mAP   & Rank-1 & Rank-5 & mAP   \\ \midrule
\multicolumn{2}{c|}{Input Size (H$\times$W)}  &\multicolumn{3}{c|}{64$\times$44}   & \multicolumn{3}{c}{128$\times$88} \\ \midrule

PoseGait~\cite{posegait}	        & PR20	 & 0.2 & 1.1 & 0.5  & -  & -  & -  \\ 
GaitGraph~\cite{gaitgraph}	& ICIP21 & 6.3  & 16.2   & 5.2	& - & -   & - \\  
GEINet ~\cite{geinet} &ICB2016  &5.4 &14.2 &5.1  &7.0 &16.3 &6.1  \\
GaitSet~\cite{gaitset}		    & AAAI19		& 36.7  & 58.3  & 30.0 & 42.6 & 63.1 & 33.7 \\
GaitPart~\cite{gaitpart}		& CVPR20		& 28.2 & 47.6 & 21.6	& 29.9 & 50.6 & 23.3 \\ 
GLN~\cite{gaitgln } 				& ECCV20		& 31.4 & 52.9 & 24.7 & 42.2 & 64.5 & 33.1	\\
GaitGL~\cite{gaitgl}	            & ICCV21		& 29.7 & 48.5 & 22.3 	& 23.5 & 38.5 & 16.4 \\ 
CSTL~\cite{cstl}	            & ICCV21		& 11.7 & 19.2 & 5.6  	& 12.2 & 21.7 & 6.4  \\ 
SMPLGait~\cite{gait3d}& CVPR22  & 46.3	& 64.5 & 37.2 & 53.2 & 71.0 & 42.4     \\
MTSGait~\cite{gait3d}& MM22  & 48.7	& 67.1 & 37.6 & - & - & -     \\  

GaitBase~\cite{gaitbase}& CVPR23   & 64.6 & 81.5 & 55.2   & - & - & - \\  

DyGait~\cite{dygait}& ICCV23   & -  & -  & -   & 66.3 & 80.8 & 56.7 \\ 
\textbf{GaitRDAE}& -       & \textbf{69.8}  & \textbf{84.8}	& \textbf{59.9}   & \textbf{73.6}  & \textbf{86.7}	& \textbf{63.5} \\

\bottomrule
\end{tabular} 
\label{tab:gait3d}
\end{table*}

\begin{table}[t]
  \centering
      \renewcommand\arraystretch{1.35} 
  \setlength{\tabcolsep}{1.15mm}
  
  \caption{
  Performance comparison of Rank-1 (\%), Rank-5 (\%), Rank-10 (\%), and Rank-20 (\%) on GREW.
  }
\resizebox{0.5\textwidth}{!}{
\begin{tabular}{l|c|cccc}
\toprule
Method & Source & Rank-1 & Rank-5 & Rank-10 & Rank-20 \\
\midrule
    PoseGait~\cite{posegait}& PR20& 0.2  & 1.1  & 2.2  & 4.8  \\ 
    GaitGraph~\cite{gaitgraph}& ICIP21& 1.3  & 3.5  & 5.1  & 7.5  \\
    GEINet~\cite{geinet} &ICB16 & 6.8   & 13.4  & 17.0  & 21.0      \\
    TS-CNN~\cite{gei3}& TPAMI16& 13.6  & 24.6  & 30.2  & 37.0  \\ 
    GaitSet~\cite{gaitset} & AAAI19& 46.3  & 63.6  & 70.3  & 76.8  \\
    GaitPart~\cite{gaitpart}& CVPR20& 44.0  & 60.7  & 67.3  & 73.5  \\
    GaitGL \cite{gaitgl} &ICCV21 & 47.3  & 63.6  & 69.3  &74.2  \\ 
    CSTL\cite{cstl} &ICCV21  &50.6 &65.9 &71.9 &76.9\\
    MTSGait\cite{MTSgait} &MM22 &55.3 &71.3 &76.9 &81.6 \\
    GaitBase \cite{gaitbase} &CVPR23 & 60.1  & 75.7  & 80.5  & 84.4  \\ 
    DyGait \cite{dygait} &ICCV23 & 71.4  & 83.2  & 86.8  &89.5  \\
    \textbf{GaitRDAE}  & --&\textbf{78.6} & \textbf{88.3} & \textbf{91.1} & \textbf{93.0} \\  
       \bottomrule
\end{tabular}
}
\label{tab:grew}
\end{table}

\subsection{Implementation Details}
We employ SGD\cite{SGD} for optimization, with an initial learning rate of 0.1 and a weight decay factor of 0.0005.
Each frame is standardized following the method described in \cite{ou-mvlp}, and resized to dimensions of 64 × 44 or 128 × 88.
For the training stage, the length of gait sequences is standardized to 30 frames. Conversely, for evaluation during the testing stage, the complete gait sequence is processed by our model.
In our GaitRDAE structure, the inaugural convolutional layer in the first phase employs 3D convolutions with a kernel dimension of 3$\times$3$\times$3, while subsequent layers, in parallel with RDA, use 2D convolutions with a kernel dimension of 3$\times$3.
The implementation specifics for the datasets CASIA-B, OU-MVLP, Gait3D, and GREW are as follows:
Batch sizes are configured to (8, 8), (32, 8), (32, 4), and (32, 4) for each dataset respectively.
\textcolor{black}{
The learning rate scheduler is set to step down at [20K, 40K, 50K], [60K, 80K, 100K], [20K, 40K, 50K], [50K, 100K, 150K] iterations respectively for each dataset.
}
The training is set to proceed for 60k, 120k, 160k, and 60k iterations for each respective dataset.
For CASIA-B, the number of output channels at each stage in Figure~\ref{overview} is [64, 128, 256, 256], and for the remaining datasets, it is [64, 128, 256, 512].
The margin $m$ in eq.\ref{tri} is assigned a value of 0.2.
In the CME, the local temporal average pooling utilizes a kernel of size 5 with a stride of 1 and padding of 2.
Codebase OpenGait\cite{gaitbase} is used for all experimental methods, which are implemented using PyTorch\cite{pytorch}.

\subsection{Comparison with the State-of-the-art Methods}
\subsubsection{Evaluation Results for in-the-wild Scenarios}
GREW and Gait3D are two datasets collected under unconstrained conditions in recent years and are receiving increasing research attention because of their strong relevance to the practicality of the methods. The gait data collected in these datasets incorporates the effects of many broad and complex covariates (e.g., background, clothing, posture) in the real world.

\textbf{GREW.} 
In Table \ref{tab:grew}, a performance comparison of our GaitRDAE methodology against leading techniques demonstrates our superior accuracy across various ranks from the GREW dataset. 
For example, our method outperforms DyGait by substantial margins of 7.2\% in Rank-1, 5.1\% in Rank-5, 4.3\% in Rank-10, and 3.5\% in Rank-20 metrics, respectively.
Furthermore, when compared to GaitBase, our method achieves impressive improvements, surpassing it by 18.5\% in Rank-1, 12.6\% in Rank-5, 10.6\% in Rank-10, and 8.6\% in Rank-20 metrics, respectively.
The outcomes of the experiment demonstrate that our approach's practicality and robustness in dealing with complex covariates in the real world, proving its effectiveness as a state-of-the-art solution.

\textbf{Gait3D.}
Table \ref{tab:gait3d} provides additional evidence of the superiority
 of GaitRDAE by comparing our performance with the latest techniques on the Gait3D dataset, specifically in terms of Rank-1, Rank-5, and mAP scores. 
When evaluated at 128x88 resolution, our approach demonstrates superior performance over DyGait, exhibiting improvements of 7.3\%, 5.9\%, and 6.8\% in Rank-1, Rank-5, and mAP\cite{map-hou2022comprehensive} metrics, respectively. When evaluated at 64$\times$44 resolution, our technique surpasses GaitBase, achieving gains of 
5.2\%, 3.3\% and 4.7\%
in Rank-1, Rank-5, and mAP metrics, respectively. 
These results clearly demonstrate the robust discriminative capabilities of our method, especially in terms of gait recognition in the face of uncertainties present in real-world 
scenarios.
\textcolor{black}{
The superior performance at higher resolutions can be attributed to enhanced detail and clarity, improved feature extraction by the RDA  and RDE modules. Additionally, higher resolution inputs help reduce noise and make better use of the model's capacity as these inputs provide finer granularity and clearer representations, enabling more precise adjustments and robust feature learning, which are critical for gait recognition.
}
\subsubsection{Evaluation Results for in-the-lab Scenarios}
CASIA-B and OU-MVLP are two classical laboratory gait datasets collected in a constrained environment in earlier years and used by most of the previous methods to validate the validity of the method.

\begin{table}[t]
  \centering
  \small
  \renewcommand\arraystretch{1.4} 
  \setlength{\tabcolsep}{1.2mm}  
\caption{Performance comparison of Rank-1 (\%) performance on CASIA-B. Cases with identical-views are excluded. Bold numbers indicate the best results and underlined numbers represent the second-best results.}
\begin{threeparttable}
\resizebox{0.4\textwidth}{!}{
\begin{tabular}{l|c|cccc}
\toprule
Method & Size & NM & BG &CL & Mean \\
\midrule
GaitSet~\cite{gaitset} & $64 \times 44 $   &  95.0  &   87.2 &  70.4  &84.2                \\ 
GaitPart~\cite{gaitpart}&$64 \times 44 $       &   96.2   & 91.5   &  78.7  & 88.8                   \\
MT3D~\cite{mt3d}        &$ 64 \times 44 $   &96.7    & 93.0   & 81.5   &  90.4            \\
CSTL~\cite{cstl}        &  $64 \times 44 $  &  97.8  & 93.6   &  84.2  &  91.9            \\
3DLocal~\cite{3dlocal}        & $64 \times 44$     &  97.5  &94.3    & 83.7   &  91.8             \\
GaitGL~\cite{gaitgl}        & $ 64 \times 44 $   & 97.4   &  94.5  &  83.6  &  91.8              \\
L-Gait ~\cite{lagrangegait}        & $64 \times 44 $   &   96.9    &  93.5  &   86.5   &   92.3            \\ 
GaitBase~\cite{gaitbase}        & $64 \times 44 $  &  97.6     &  94.0        & 77.4   &   89.7         \\ 
DyGait~\cite{dygait}        & $64 \times 44$    &  \textbf{98.4}     &  \textbf{96.2}        & 87.8   &  94.1          \\ 
\textbf{GaitRDAE}~   & $64 \times 44  $   & \underline{97.8}    &  \underline{95.9}  &   \textbf{88.2}   &      \underline{94.0}       \\ 
\hline
GLN~\cite{gaitgln}        & $128\times 88 $   & 96.9   &  94.0  &  77.5  &  89.5             \\
3DLocal~\cite{3dlocal}        & $128\times 88 $   & 98.3   &  95.5  &  84.5  &  92.7             \\
CSTL~\cite{cstl}        & $128\times 88 $   & 98.0   &  95.4  &  87.0  &  93.5             \\

\textbf{GaitRDAE}~   & $128\times 88 $    & \textbf{98.7}    &   \textbf{97.2}   &   \textbf{90.6}   &      \textbf{95.5}       \\ 
       \bottomrule
\end{tabular}
}

  \end{threeparttable}
  
\label{tab:CASIA-B}
\end{table}

\textbf{CASIA-B.} 
Table \ref{tab:CASIA-B} presents a comparative evaluation of our GaitRDAE approach against the latest advancements, focusing on average Rank-1 accuracy metrics from the CASIA-B dataset. At an input resolution of 128x88, our technique outperforms the others with an average Rank-1 accuracy of 95.5\% in the NM, BG, and CL conditions. This reflects a performance improvement of 1.4\% over CSTL and 5.8\% over 3DLocal. 
At a reduced resolution of 64$\times$44, our method holds a competitive position, ranking second in average retrieval accuracy. Remarkably, in the challenging CL state, where pedestrian appearances vary significantly, our method demonstrates superior accuracy at both resolutions. For 128x88 resolution, it surpasses CSTL by 3.6\% and 3DLocal by 6.1\%; at 64x44, it exceeds DyGait by 0.4\%. These achievements underscore the robustness of our approach and its superiority  in extracting dynamic patterns that are independent of appearance.

\textbf{OU-MVLP.} 
Table \ref{tab:OUMVLP} offers a comparative evaluation of our GaitRDAE approach against existing state-of-the-art methods, focusing on the averaged Rank-1 accuracy metrics derived from the OU-MVLP dataset.
From the data presented in Table~\ref{tab:OUMVLP}, several key points stand out:
For average accuracy across all viewpoints, GaitRDAE achieves optimal average accuracy compared to previous methods;
For recognition accuracy under all viewpoints, GaitRDAE achieves optimal results under the majority of viewpoints.
It is noteworthy that GaitRDAE is effective in improving accuracy in viewpoints with limited dynamic information, such as 0 and 180 degrees. This phenomenon can be attributed to GaitRDAE's ability to effectively mine discriminative motion regions from which robust features can be extracted.

\begin{table*}[t]
\centering
\small
\renewcommand{\arraystretch}{1.4}
\setlength{\tabcolsep}{1.35mm}
\caption{Performance comparison of Rank-1 (\%) on OU-MVLP. Cases with identical-views are excluded.}
\begin{tabular}{c|c|cccccccccccccc|c}
\toprule
\multirow{2}{*}{Method} & \multicolumn{1}{c|}{\multirow{2}{*}{Source}} & \multicolumn{14}{c|}{Probe View}                                                               & \multirow{2}{*}{Mean} \\ \cmidrule{3-16}
                        & \multicolumn{1}{c|}{}                      & 0$^{\circ}$ & 15$^{\circ}$ & 30$^{\circ}$ & 45$^{\circ}$ & 60$^{\circ}$ & 75$^{\circ}$ & 90$^{\circ}$ & 180$^{\circ}$ & 195$^{\circ}$ & 210$^{\circ}$ & 225$^{\circ}$ & 240$^{\circ}$ & 255$^{\circ}$ & \multicolumn{1}{c|}{270$^{\circ}$} &      \\ \midrule
                                            GaitSet~\cite{gaitset}                                      & AAAI19  &79.5&87.9&89.9&90.2&88.1&88.7&87.8&81.7&86.7&89.0&89.3&87.2&87.8&86.2&87.1           \\
                      GaitPart~\cite{gaitpart}   &CVPR20                                  &   82.6&88.9&90.8&91.0&89.7&89.9&89.5&85.2&88.1&90.0&90.1&89.0&89.1&88.2&88.7    \\
                    GLN~\cite{gaitgln}           &ECCV20                               &  83.8&90.0&91.0&91.2&90.3&90.0&89.4&85.3&89.1&90.5&90.6&89.6&89.3&88.5&89.2         \\
                    GaitKMM\cite{gaitkmm} &CVPR21 & 56.2  & 73.7 & 81.4 & 82.0 & 78.4 & 78.0 & 76.5 & 60.2 & 72.0 & 79.8 & 80.2 & 76.7 & 76.3 & 73.9 & 74.7 \\
                    CSTL~\cite{cstl}              &ICCV21                           &   87.1&91.0&91.5&91.8&90.6&90.8&90.6&89.4&90.2&90.5&90.7&89.8&90.0&89.4&90.2           \\
                      3DLocal~\cite{3dlocal}       &ICCV21                               &   86.1&91.2&\textbf{92.6}&\textbf{92.9}&92.2&91.3&91.1&86.9&90.8&\textbf{92.2}&92.3&91.3&91.1&90.2&90.9        \\
                     GaitGL~\cite{gaitgl}        &ICCV21                               &  84.9&90.2&91.1&91.5&91.1&90.8&90.3&88.5&88.6&90.3&90.4&89.6&89.5&88.8&89.7       \\
                    L-Gait~\cite{lagrangegait}   &  CVPR22                                          & 85.9  & 90.6    & 91.3   & 91.5   & 91.2   & 91.0   &90.6    &88.9     &89.2     &90.5     &90.6     & 89.9    & 89.8    &     89.2                     & 90.0    \\ 
                     
                     GaitBase~\cite{gaitbase}   &  CVPR23                                       & -  & -    & -   & -   & -   & -   & -    & -    & -    & -     & -     & -  & -    & -                     & 90.8    \\ 
                       \textbf{GaitRDAE}   &  -                                          & \textbf{89.8}  & \textbf{92.0}    &92.0   & 92.2   & \textbf{92.3}   & \textbf{91.8}   &\textbf{91.7}    &\textbf{91.5}     &\textbf{91.4}     &91.5     &\textbf{91.6}     & \textbf{91.5}    & \textbf{91.1}    &     \textbf{90.9}                     &\textbf{91.5}     \\ 
                        \bottomrule

\end{tabular}
 \label{tab:OUMVLP}
\end{table*}

\subsection{Ablation Study}
To evaluate the validity of Region-aware Dynamic Aggregation (RDA) and Region-aware Dynamic Excitation (RDE) within our framework in real-world settings, the ablation studies are conducted on the large-scale unconstrained gait dataset, Gait3D. 
It should be highlighted that our baseline in the first row of   Table\ref{baseline} does not incorporate any of the modules proposed in this paper.

\textbf{Effectiveness of RDA and RDE.}
As evident from the data presented in Table\ref{baseline}, the subsequent conclusions can be established:
The integration of RDA yields substantial enhancements, resulting in a remarkable 6.3\% increase in Rank-1 accuracy and a 8.0\% boost in mAP, which highlights the importance of applying region-aware dynamic temporal patterns;
The introduction of RDE 
results in an enhancement of 1.0\% in both Rank-1 accuracy and mAP,
which confirms the validity of region-aware dynamic excitation.
The combination of  RDA and RDE produces more significant improvements, with a noteworthy 7.6\% increase in Rank-1 accuracy and an impressive 8.8\% boost in mAP.
These findings illustrate the complementary roles of RDA and RDE, i.e., RDE excitates the key motion regions and provides RDA with the necessary regional information prior to searching for the optimal temporal receptive field for each region, highlighting their combined use as a synergistic approach.
\begin{table}[t]
\centering
\renewcommand\arraystretch{1.2} 
\caption{
Ablation investigation on the effectiveness of RDA and RDE on Gait3D.
}
\label{baseline}
\resizebox{0.38 \textwidth}{!}{
\begin{tabular}{cc|ccc}
\toprule
\multicolumn{1}{c}{RDA} & RDE & \multicolumn{1}{c}{Rank-1} & Rank-5 & mAP   \\ \midrule
\multicolumn{1}{c}{}   &     & 62.2 &77.6  & 51.1    \\ 
\multicolumn{1}{c}{\checkmark}    &    & 68.5 &84.2 &59.1     \\ 
\multicolumn{1}{c}{}  &\checkmark  &63.2 &78.1  &52.1     \\ 
 \multicolumn{1}{c}{\checkmark}   & \checkmark   & \textbf{69.8} &\textbf{84.8}  &\textbf{59.9}     \\ \bottomrule
\end{tabular}
}
\end{table}

\textbf{Analysis of RDA module.} 
To investigate the suitable channel offset proportion within the RDA component, we structured an ablation study using varied offset proportions. 
Our experimental results, as presented in Table~\ref{RDAab}, lead to these conclusions:
When the channel offset ratio is increased from $\frac{1}{8}$ to $\frac{1}{4}$, the performance is improved.
This finding indicates that a higher channel offset ratio positively influences the model's performance.
When  the channel offset ratio is further increased  from $\frac{1}{4}$ to $\frac{1}{2}$, the performance degrades.
This phenomenon may be due to the gradual increase in the channel offset ratio, which may affect the spatial features of the model, which is consistent with previous findings\cite{TSM}.
\textcolor{black}{
Consequently, in this study, we choose to set the channel offset ratio at $\textcolor{black}{r}=\frac{1}{4}$,
this choice is made to strike a  balance between augmenting temporal and spatial modeling capabilities, ensuring an optimal trade-off for our proposed approach.
}

\begin{table}[t]
\caption{\textcolor{black}{Ablation investigation on the channel offset ratio \(r\) of the RDA module on Gait3D. \(r=0\) means that the model does not contain RDA, and when \(r>0\), for example, \(r = \frac{1}{4}\), it means that \(\frac{1}{4}\) of the channels are utilized to perform region-aware dynamic aggregation to achieve higher performance, which is adopted as our default setting in our experiment.}}
\centering
\label{RDAab}
\renewcommand\arraystretch{1.2}
\resizebox{0.3 \textwidth}{!}{
\begin{tabular}{c|cccc}
\toprule
$\textcolor{black}{r}$          & Rank-1 & Rank-5  &mAP  \\ \midrule
0          & 63.2 &78.1 & 52.1  \\ 
$\frac{1}{8}$          &68.7   &84.2       &59.6   \\ 
$\frac{1}{4}$          & \textbf{69.8} &\textbf{84.8}  &\textbf{59.9}  \\ 
$\frac{1}{2}$          & 69.4  & 83.5 & 58.6        \\ 
 \bottomrule
\end{tabular}
}
\end{table}

\textbf{Analysis of RDE module.} 
To investigate the impact of Spatial-wise Motion Excitation (SME) and Channel-wise Motion Excitation (CME), we conducted ablation studies examining the significance of these components. \textcolor{black}{And also, to obtain the insights on the distinct property of these two modules, we provided the comparison performances with other two well-known spatial and attention methods, Spatial Attention Module (SAM) \cite{CBAM} and Squeeze-and-Excitation Networks (SENet)\cite{SENet}. } 
As outlined in Table~\ref{RDEab}, several observations come to light:
SME and CME both yield improvements in recognition performance, which underscore the benefit of the enhancement of dynamic information across spatial and channel dimensions; \textcolor{black}{Moreover, the superior performances of SME and CME comparing with the corresponding SAM and SENet demonstrate that different from conventional image-based task, gait recognition is a temporal sequential task. Thus, enhancement in the spatial/channel dimensions associated with temporal dynamic is essential to extract discriminative gait pattern under complex covariate conditions.}
Combining the SME and CME techniques together maximized the performance improvement, resulting in a 1.4\% improvement in Rank-1 accuracy, which supports the idea that enhancing dynamic information from both spatial and channel dimensions is not redundant but complementary, thus further improving the recognition capability of our proposed approach.

\begin{table}[t]
\centering
\caption{
Ablation investigation on the component of RDE on Gait3D.
}
\renewcommand\arraystretch{1.25}
\label{RDEab}
\resizebox{0.45  \textwidth}{!}{
\begin{tabular}{cccc|ccc}
\toprule
SME & \textcolor{black}{SAM} & CME & \textcolor{black}{SENet} & \multicolumn{1}{c}{Rank-1} &Rank-5 &mAP  \\ \midrule
\multicolumn{1}{c}{}   &  & &  & 68.5 &84.2 &59.1  \\ 
\multicolumn{1}{c}{\checkmark} &  &  &   &69.1 &84.8  &59.4   \\ 
 & \textcolor{black}{\checkmark} &  &   & \textcolor{black}{68.4} & \textcolor{black}{84.0}  & \textcolor{black}{59.3}   \\
  &  &  \checkmark &  & 68.8 & 84.7  & 59.4   \\
   &  &  & \textcolor{black}{\checkmark}  & \textcolor{black}{68.3} & \textcolor{black}{82.3}  & \textcolor{black}{58.2}   \\
 \multicolumn{1}{c}{\checkmark} &  & \checkmark &  & \textbf{69.8} &\textbf{84.8}  &\textbf{59.9}   \\ \bottomrule
\end{tabular}
}
\end{table}

\subsection{Visualization}
\subsubsection{Adaptive Temporal Scale Visualization}
\begin{figure}[t]
\begin{center}
\includegraphics[width=1.0\linewidth]{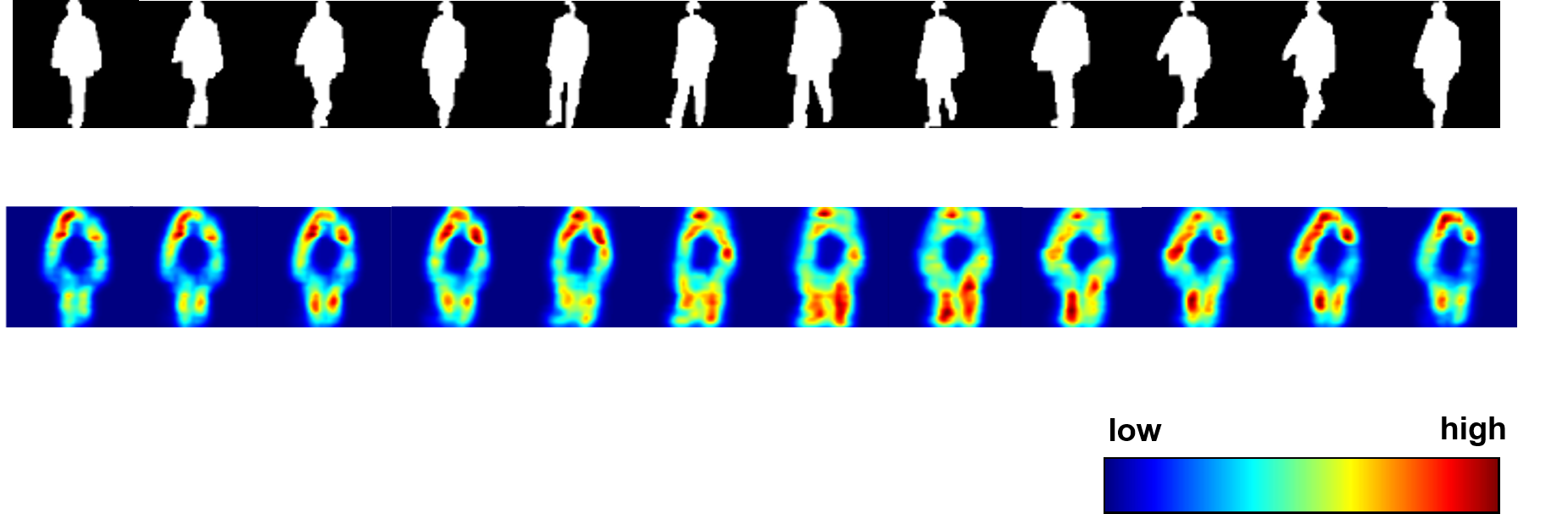}
\end{center}
     \caption{
  Visualization of the learnable temporal offset value. 
   }
   \label{heat}
   \end{figure}

Figure~\ref{heat}  depicts a visualization of the learnable temporal offset value $\Delta$$p$  given a sequence of gait silhouettes. The red regions represent pixels assigned large temporal offset values, and the blue regions are the opposite. As the visualization shows, our method can automatically search for motion regions (e.g., hands, feet) and assign longer temporal receptive fields to them while assigning restricted temporal receptive fields to static regions. \textcolor{black}{It also demonstrates that the learned offset values naturally cluster into coherent regions. This is because pixels within the same semantic region share similar motion patterns, leading the model to assign them similar temporal scales through the learning process.}

\subsubsection{Visualization of Feature Distributions}
\begin{figure}[t]
\begin{center}
\includegraphics[width=1.0\linewidth]{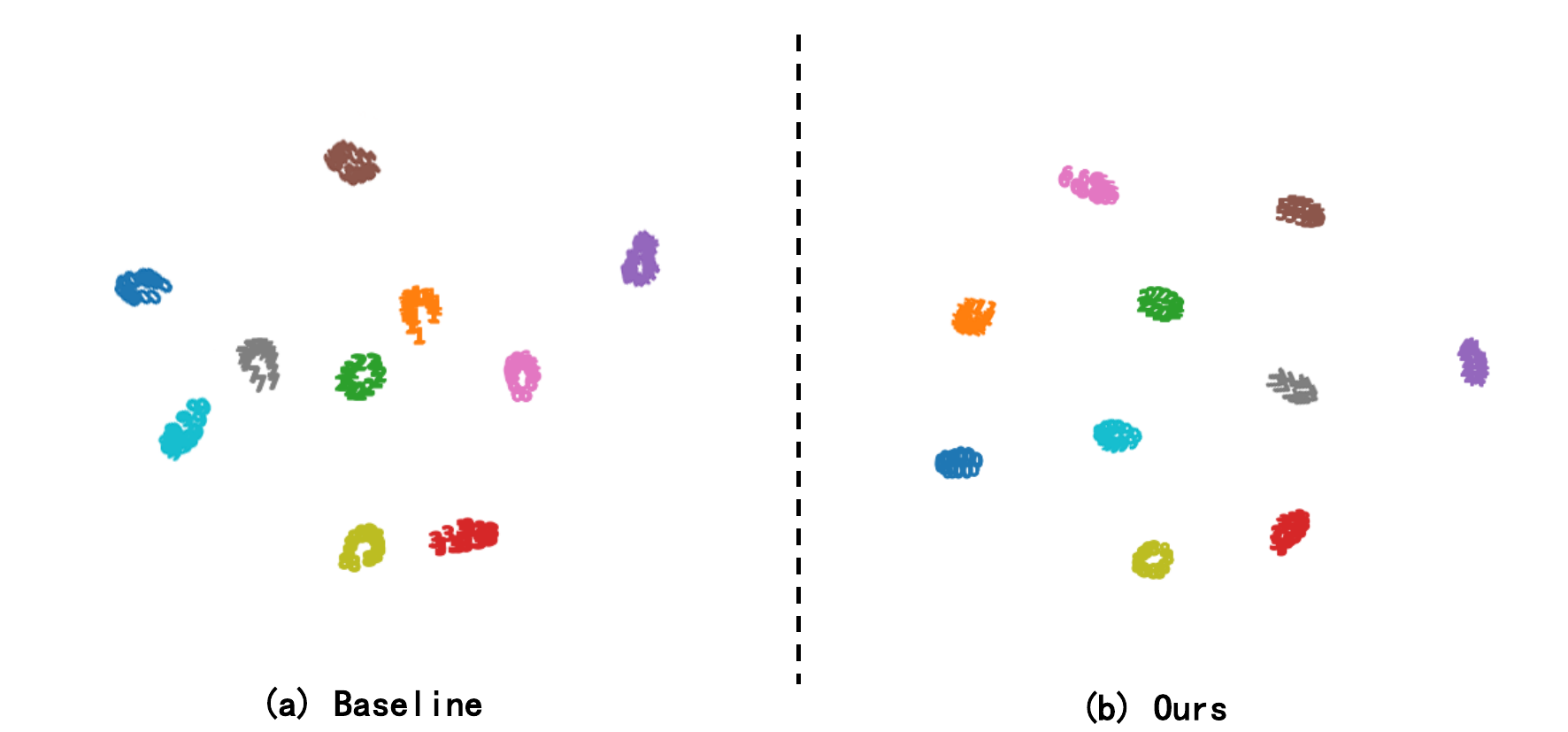}
\end{center}
   \caption{
Feature distribution visualization is achieved using t-SNE\cite{tsne}:
(a) Baseline's feature distribution.
(b) GaitRDAE's feature distribution.
}
   \label{tsne}
\end{figure}

Figure \ref{tsne} illustrates the feature distribution for ten distinct identities from the CASIA-B dataset, selected at random. The distribution is visualized using t-SNE \cite{tsne} to evaluate the performance of our GaitRDAE method. The figure reveals that GaitRDAE achieves a more compact 
intra-class 
feature distribution and more expansive
inter-class
separation than the Baseline method. This indicates that GaitRDAE is capable of deriving more discriminative features, which are crucial for accurately differentiating between the gait patterns of various individuals.

\begin{figure}[t]
\begin{center}
\includegraphics[width=1.0\linewidth]
{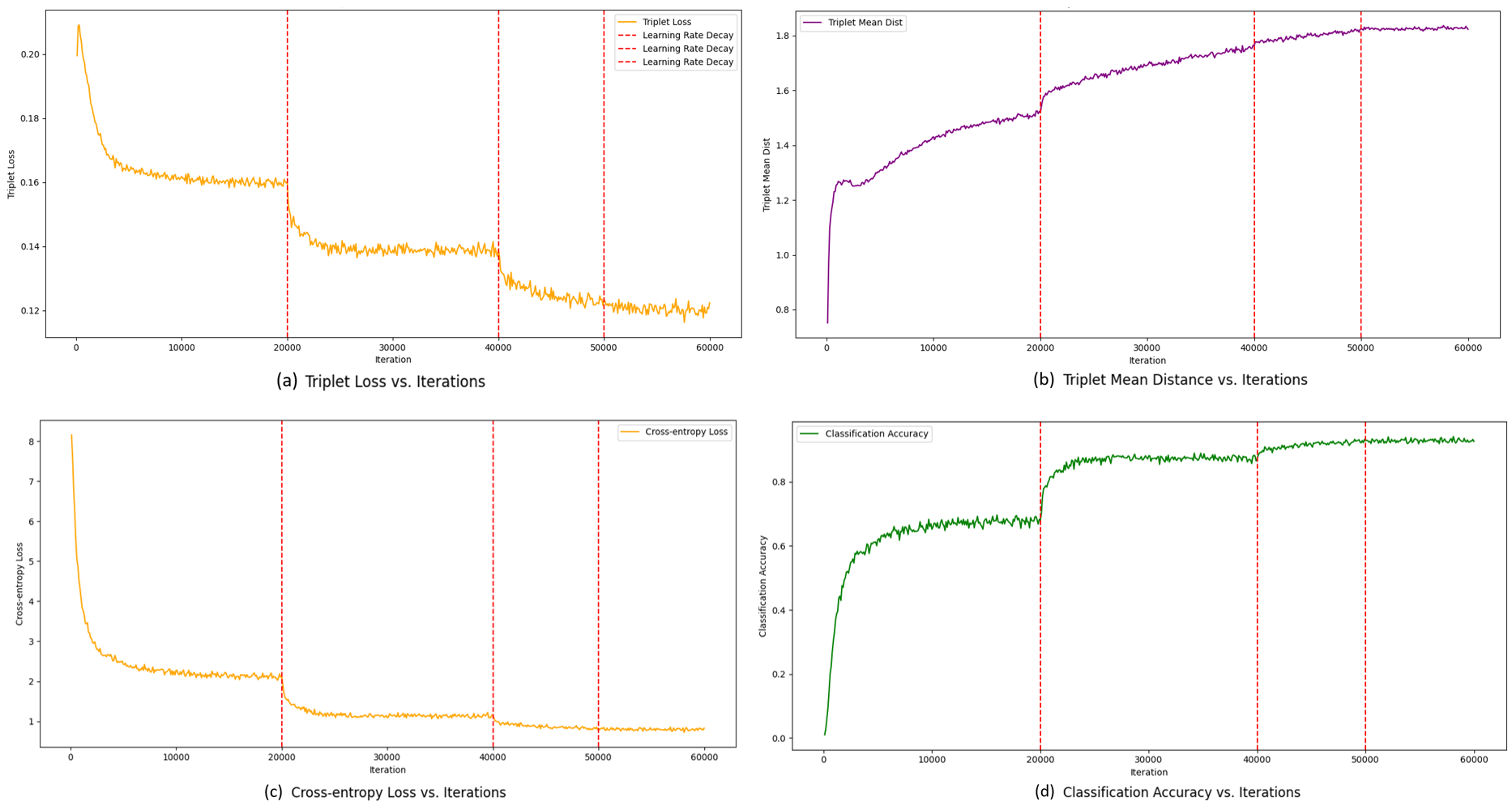}
\end{center}
   \caption{
\textcolor{black}{Analysis of loss components and accuracy over training iterations on Gait3D. (a) Triplet Loss vs. Iterations, (b) Triplet Mean Distance vs. Iterations, (c) Cross-Entropy Loss vs. Iterations, and (d) Classification Accuracy vs. Iterations. The vertical red dashed lines indicate learning rate decay points at 20K, 40K, and 50K iterations. }
   }
   \label{combined_loss}
\end{figure}

\textcolor{black}{
\subsection{Analysis of Loss Components}
In this section, we conducted a detailed analysis of the loss components during the training process, focusing on triplet loss, triplet mean distance, cross-entropy loss, and classification accuracy over the training iterations. As shown in Fig. \ref{combined_loss}, the triplet loss gradually decreases, indicating effective minimization of intra-class distances and the overall optimization of triplet relationships. Simultaneously, the triplet mean distance increases, showing enhanced inter-class separability and the model's ability to differentiate between distinct classes. Learning rate decays at 20K, 40K, and 50K iterations significantly contribute to these improvements, as indicated by the sharp changes in the trends. The cross-entropy loss decreases substantially, reflecting improved classification performance, and the classification accuracy consistently improves, with noticeable jumps at the learning rate decay points, confirming the positive impact of learning rate adjustments on model optimization. Overall, the strategic application of learning rate decay accelerates convergence and boosts model performance. These trends validate the model's capability to enhance intra-class compactness and inter-class separability, demonstrating its robust classification performance and overall efficacy in gait recognition tasks.
}

\textcolor{black}{
\section{Limitations and Future Work}
The physiological structure of the human body dictates that different regions exhibit distinct motion characteristics during walking. Therefore, enabling a gait model to capture unique motion features of different body parts and allocate appropriate temporal receptive fields for these regions is important for recognition. Our framework addresses this challenge by proposing an adaptive aggregation and excitation approach. 
We acknowledge that our current method leverages 3D convolutions to search for receptive fields across different regions and capture spatiotemporal dependencies. While effective, this approach results in an increase in the number of parameters. 
Looking forward, we plan to explore the use of model distillation and compression techniques, as well as 2D+1D decomposed modeling approaches, to more efficiently model spatiotemporal dependencies. Additionally, we aim to integrate the latest parsing-based methods to enhance the understanding and modeling of motion regions. These advancements are expected to further optimize the model's performance and efficiency.
We believe that improving the gait model's adaptive perception of different human motion regions will significantly enhance its performance in complex environments, bringing us closer to practical applications of gait recognition.
}

\section{Conclusion}
We introduced a novel framework for gait recognition, termed GaitRDAE, that enhances the robustness of gait features by learning region-aware temporal representations as well as enhancing the perception of motion regions. Specifically, Region-aware Dynamic Aggregation adaptively adjusts the receptive fields of different regions according to the input to fully adapt to the motion characteristics of different regions. Furthermore, Region-aware Dynamic Excitation enhances perceptual learning of motion regions to fully extract discriminative behavior patterns. 
Finally,  experiments on GREW, Gait3D,  CASIA-B, and OUMVLP datasets containing a variety of scenes fully conclusively validates the robustness of our model. 
By introducing these innovative region-aware learning techniques, we believe that our framework has the potential to propel the field of gait recognition forward, offering improved performance in real-world scenarios.

{
\bibliographystyle{IEEEtran}
\bibliography{main}
}

\end{document}